\RequirePackage[T1]{fontenc}
\documentclass[10pt,conference]{IEEEtran}
\IEEEoverridecommandlockouts
\usepackage[nocompress]{cite}

\usepackage{amsmath,amssymb}
\usepackage{graphicx}
\usepackage{booktabs,array,tabularx}
\newsavebox{\satopstablebox}
\usepackage{longtable}
\usepackage[hyphens]{url}
\expandafter\def\expandafter\UrlBreaks\expandafter{\UrlBreaks\do\0\do\1\do\2\do\3\do\4\do\5\do\6\do\7\do\8\do\9}
\usepackage{balance}
\usepackage[hidelinks]{hyperref}
\hypersetup{pdfauthor={Lifan Mei, Yuxuan Li, Jiaxian Tu, Shenghe Xu},pdftitle={SatOps-Decision: Low-Latency Discriminative Decisions for Onboard Satellite Communication Operations}}

\title{SatOps-Decision: Low-Latency\\Discriminative Decisions for Onboard\\Satellite Communication Operations}
\author{
  \IEEEauthorblockN{Lifan Mei\textsuperscript{1,*}, Yuxuan Li\textsuperscript{2},
    Jiaxian Tu\textsuperscript{3}, Shenghe Xu\textsuperscript{4}}
  \IEEEauthorblockA{\textsuperscript{1,2,3}Academy of AIAT, Xi'an Jiaotong-Liverpool University; \textsuperscript{4}Amazon, Inc.\\
    Email: \textsuperscript{1}lifan.mei@xjtlu.edu.cn, \textsuperscript{2}Yuxuan.Li2302@student.xjtlu.edu.cn\\
    \textsuperscript{3}Jiaxian.Tu23@student.xjtlu.edu.cn, \textsuperscript{4}shenghexu@gmail.com}
  \thanks{Shenghe Xu's work was performed outside his role at Amazon.}
  \thanks{\textsuperscript{*}Corresponding author: Lifan Mei.}
}
\IEEEaftertitletext{\vspace{-36pt}}
\renewcommand{\footnoterule}{%
  \kern-3pt\hrule width .4\columnwidth height .4pt\kern2.6pt}

\long\def\SatOpsManuscript{%
\maketitle
\bstctlcite{satops:reference-control}
\thispagestyle{empty}
\pagestyle{empty}

\begin{abstract}
We propose SatOps-Decision, a method for low-latency decisions in satellite communication operations. To the best of our knowledge, it is the first application of Jev-style general-purpose discriminative decision models to satellite operations, and the accompanying benchmark is the first dedicated to this approach. The learned scorer, a locally adapted Laya encoder with its native decision head, selects among supplied operational answers in a single forward pass. The benchmark combines authored operational contracts, controlled state changes and independently checked labels in an auditable construction. Evaluation prioritizes complete-request latency and raw decision accuracy; all main accuracy comparisons use the same 1,024 held-out questions, separate from training and development. SatOps-Decision's local scorer has a median latency of 90.3 ms per decision, versus 493.9 ms for Qwen3.5-0.8B generation and 388.4 ms for OneJev1.6-0.8B: 5.47$\times$ and 4.30$\times$ speedups based on median request times in the measured FP32/MPS implementations. Its mean raw accuracy is 85.3\%, versus 42.5\% for frozen Laya, 40.7\% for Qwen generation and 58.6\% for OneJev. As cloud accuracy references, original Jev 1.14/1.13 reach 87.4/81.9\%.
\end{abstract}


\section{Introduction}
Satellites provide essential infrastructure for global communications, navigation and Earth observation. Advances in artificial intelligence (AI) are extending their role from data collection and relay to onboard inference and orbital edge computing~\cite{orbitaledge,phisat2}. As these applications expand, demand for computing-capable satellites and onboard processing is expected to grow. Efficient communication operations are therefore central to sustaining these services under limited contact time and onboard resources.

Satellite communication operations require timely judgments about incomplete or aging evidence, available contact time and limited onboard resources. A link outage can indicate an expected contact gap or a fault requiring intervention. A backup path may exist but remain unusable because energy, readiness or completion time is insufficient. Even after a command is issued, recovery must be confirmed from subsequent observations. Delay/disruption-tolerant networking supports data retention across contact interruptions~\cite{nasadtn}; operational diagnosis, action selection and recovery verification require complementary decision mechanisms.

We propose \textbf{SatOps-Decision}, a method for low-latency decisions in satellite communication operations. Its learned scorer directly selects among supplied operational answers, while separate deterministic checks enforce execution prerequisites. The scorer receives the policy, state, question and candidate meanings through the general-purpose discriminative interface exemplified by Jev~\cite{jevlaunch,jev}. Domain-adapted Laya~\cite{laya}, a third-party encoder-based model with a native decision head, implements this scorer locally. Fig.~\ref{fig:communication-scene} shows the intended setting and the four judgments made by the scorer: evidence assessment, action selection, switching permission and recovery verification.

Language models have been explored as spacecraft operators and satellite agents~\cite{spacecraftoperators,satelliteopsgym}. When an operational question has a small set of supplied answers, generating a textual answer adds an answer-decoding stage that direct scoring can avoid. This makes short request latency the main motivation for our method. We ask whether domain adaptation improves decision accuracy while retaining that computational advantage. We test SatOps-Decision's domain-adapted Laya scorer against frozen Laya, local generative Qwen~\cite{qwen}, local OneJev~\cite{onejev} and original Jev accessed through its cloud API.

Evaluating such a component also requires a benchmark that tests the decisions themselves. Switching permission depends on operational prerequisites, and recovery requires observations confirming the command's effects. We therefore define an explicit operational contract with independently checked targets and controlled changes in evidence, contact and resource conditions. Every model receives the same policy, state and candidate meanings. The construction fits the Jev-style interface.

Our contributions are:
\begin{enumerate}
\item \textbf{The first application of Jev-style models to satellite operations.} To the best of our knowledge, we are the first to apply Jev-style general-purpose discriminative decision models to satellite operations. SatOps-Decision instantiates this approach for communication operations with a domain-adapted local Laya scorer, evaluated primarily by complete-request latency and raw held-out decision accuracy.
\item \textbf{The first dedicated Jev-style satellite operations benchmark.} To the best of our knowledge, this is the first benchmark designed specifically for Jev-style satellite operations. It provides fully specified operational contracts, a reproducible construction procedure and an auditable evaluation interface shared by all models, with separate training, development and test data.
\end{enumerate}

\begin{figure}[t]
\centering
\vspace*{4pt}
\includegraphics[width=.97\columnwidth]{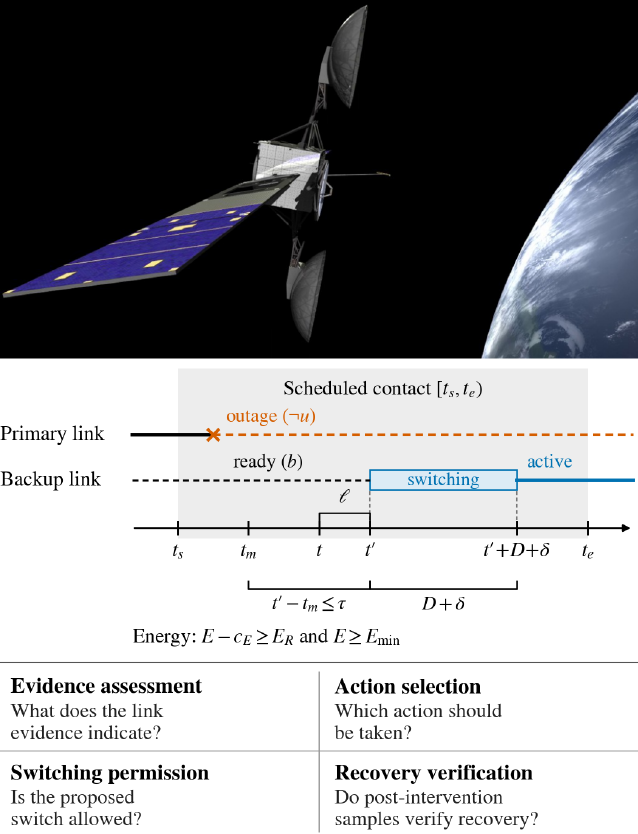}
\caption{Satellite communication setting and four operational judgments. Evidence from $t_m$ is assessed at $t$, with an answer at $t'=t+\ell$. Switching must fit within contact $[t_s,t_e)$, reserve $D+\delta$ for the switch and margin, and meet the freshness and energy conditions in~\eqref{eq:permit}. Times are in seconds; energy uses synthetic percentage points. TDRS artwork: NASA/Goddard Space Flight Center~\cite{nasatdrsart}. Time axis is not to scale.}
\label{fig:communication-scene}
\end{figure}
\section{Background and scope}
TypeSafe introduced Jev on September 15, 2026~\cite{jevlaunch}. Its supplied-state interface returns typed decisions and probabilities; Choice, Noul and Score represent categorical, binary and ordered judgments~\cite{jev}. Candidate descriptions specify what an answer means, allowing one model to serve multiple decisions without generating a free-form response. This interface supports implementations with different model backbones.

Laya is a third-party model combining a pretrained ModernBERT-large encoder with a native decision head~\cite{laya}. We adapt these pretrained components to satellite operational decisions. OneJev1.6-0.8B is derived from Qwen3.5-0.8B, which serves as our generative LLM control~\cite{onejev,qwen}. Original TypeSafe Jev weights are unavailable for local evaluation, so its 1.13/1.14 API versions provide accuracy references only.

Prior work already covers spacecraft language agents~\cite{spacecraftoperators,guide}, satellite diagnosis and recovery benchmarking~\cite{satelliteopsgym}, communication control and routing~\cite{scnoc,stragent}, and wireless System-One decisions~\cite{wirelesssystemone}. DoomSat uses ground-generated Jev decisions to control a game payload through spacecraft software~\cite{doomsat}, while Odyssey uses Jev for bounded navigation choices in a simplified cinematic flight simulation~\cite{odyssey}. These demonstrations concern game or flight-navigation choices, whereas SatOps-Decision provides a domain-adapted method and benchmark for satellite operations. RSJEV applies discriminative scoring to remote-sensing scenes~\cite{rsjev}. Xu and Mei evaluate Jev for network traffic classification, including accuracy, processing time and cost~\cite{jevtraffic}. Our benchmark builds on existing satellite operations research through a common Jev-style decision interface.

Counterfactual examples and behavioral testing are established tools~\cite{counterfactualdata,checklist}; DecisionPivot uses controlled decision groups and independent solvers~\cite{decisionpivot}. We apply these principles to contact-, resource- and evidence-conditioned operations.
\section{SatOps-Decision method}
SatOps-Decision uses a domain-adapted Laya scorer for all four operational judgments. Its Jev-style interface takes policy $r$, timestamped evidence $x$, question $q$ and candidate meanings $\mathcal A$, excluding labels, root identifiers and evaluation metadata. Fig.~\ref{fig:communication-method} shows the separate scoring and execution checks.

\begin{figure}[t]
\centering
\includegraphics[width=\columnwidth]{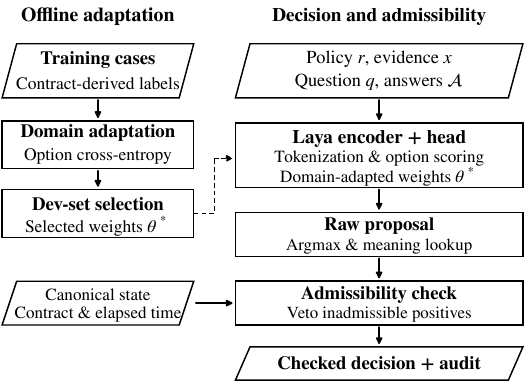}
\caption{Shared decision scorer and separate operational checks. Development data select the adapted weights; raw predictions are scored before the veto-only checker and kept separate from checked decisions.}
\label{fig:communication-method}
\end{figure}

\begin{figure*}[t]
\centering
\includegraphics[width=\textwidth]{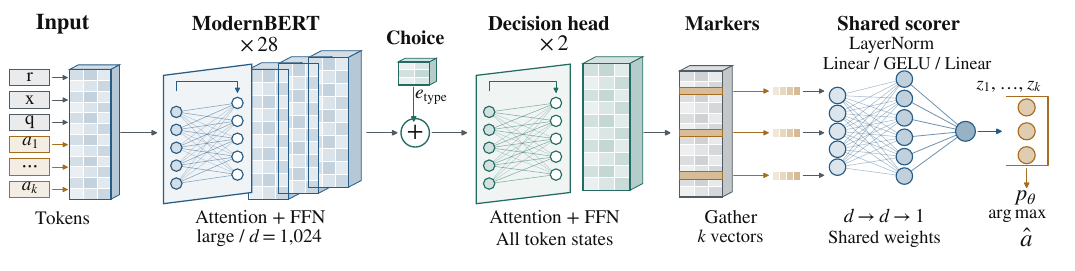}
\caption{Laya neural architecture. All tokens traverse the encoder and decision head before extracting $k=|\mathcal A|$ option vectors. Shared scoring and masked softmax yield candidate probabilities.}
\label{fig:laya-architecture}
\end{figure*}

\subsection{Local adaptation and direct answer scoring}
We adapt the pinned 421M-parameter Laya English checkpoint. Fig.~\ref{fig:laya-architecture} details the bidirectional encoder and native decision head producing option logits $z_\theta(a)$. All four tasks, including binary permission and recovery, use Choice scoring:
\begin{equation}
\begin{aligned}
 p_\theta(a\mid r,x,q,\mathcal A)&=
 \frac{\exp z_\theta(a)}{\sum_{b\in\mathcal A}\exp z_\theta(b)},\\
 \hat a&=\arg\max_{a\in\mathcal A}p_\theta(a).
\end{aligned}
\end{equation}
Padded markers are masked. A single forward pass returns candidate scores without autoregressive decoding.

The shaded modules in Fig.~\ref{fig:laya-architecture} are jointly fine-tuned using masked option cross-entropy,
\begin{equation}
 \mathcal L(\theta)=-\frac{1}{N}\sum_{i=1}^{N}
 \log p_\theta(y_i\mid r_i,x_i,q_i,\mathcal A_i).
\end{equation}
We update 421,029,889 of 421,293,827 parameters; the unused act/escalate head stays frozen. Native augmentation permutes options and remaps targets while preserving the underlying states. Optimization updates the learned scorer; the operational checker is applied after inference. Inference uses temperature 1.

Training uses 1,536 questions, four epochs and seeds 7/17/29. Checkpoints maximize raw task-macro accuracy on 256 development questions, breaking ties by earlier epoch. All three predesignated seeds are reported, with checkpoint selection completed on development data before test evaluation. The full report includes settings and ablations.

Each run uses 384 AdamW updates: microbatch 4, accumulation 4 (effective batch 16), encoder/head learning rates $2.5\times10^{-5}/10^{-4}$, weight decay 0.01, gradient-norm clip 1, and 5\% warmup then cosine decay to $10^{-6}$. Training and weights use FP32 with gradient checkpointing. Before confirmation testing, native-path checkpoint reloads must reproduce development predictions. Token budgets are 1,024 total and 256 for question/options; truncation audits preserve policy and answer meanings.

\subsection{Operational contract and execution boundary}
Let $t$ be snapshot time, $\ell$ the processing interval and $t'=t+\ell$ the time at which an answer becomes available. Evidence measured at $t_m$, with validity $v$ and maximum age $\tau$, is usable when
\begin{equation}
 F_\ell=v\wedge(0\le t-t_m)\wedge(t'-t_m\le\tau).
\end{equation}
Let $u$ denote primary availability, $b$ backup readiness, $E$ energy, $c_E$ switch cost, $E_R$ reserve, $E_{\min}$ the policy threshold, $D$ switch duration, $\delta$ its margin and $[t_s,t_e)$ scheduled contact. Switching is permissible exactly when
\begin{equation}
\begin{aligned}
 P_\ell={}&F_\ell\wedge\neg u\wedge b\wedge(t_s\le t\le t'<t_e)\\
 &\wedge(E-c_E\ge E_R)\wedge(E\ge E_{\min})\\
 &\wedge(t'+D+\delta\le t_e).
\end{aligned}
\label{eq:permit}
\end{equation}
Offline questions declare $\ell=0.5$~s; the event diagnostic substitutes the measured interval. Energy is expressed in synthetic percentage points. These authored values define the benchmark contract.

\textbf{Assessment} distinguishes insufficient evidence, normal primary operation, in-contact fault evidence and an expected gap. \textbf{Action} selects collect, monitor, switch or wait under the same evidence and feasibility priority. \textbf{Permission} separately allows or blocks switching according to~\eqref{eq:permit}. \textbf{Recovery} requires an executed intervention ending before the snapshot and at least $K$ valid, healthy observations at distinct timestamps after its completion; the latest $K$ must remain fresh at $t'$. Verification therefore uses fresh observations obtained after intervention.

The deterministic checker vetoes a proposed switch or permission that violates $P_\ell$, and a positive recovery claim lacking its required evidence. The checker preserves negative proposals. Dispatch requires consistent action and permission judgments; recovery is evaluated after new observations arrive. The primary evaluation reports the learned scorer's \emph{raw} predictions before deterministic corrections.
\section{SatOps-Decision benchmark}
\subsection{Motivation and construction}
The benchmark targets decisions that a common typed interface can express across operational stages. NASA networking and fault-response guidance motivate contact-aware interpretation and verification of observed command effects~\cite{nasadtn,nasafaultresponse}; the separation of diagnosis, intervention and recovery is inspired by SatelliteOpsGym~\cite{satelliteopsgym}. All states, policies, numerical ranges, text templates and labels are authored here; no mission telemetry or existing benchmark examples are reused.

Construction has four steps. First, a seeded generator samples an explicit policy and nominal state satisfying the relevant contract. Second, controlled changes alter evidence validity or age, contact time, backup readiness, energy or recovery observations. Third, two independently expressed contract implementations must agree on every label. Finally, question wording and answer-slot mappings are varied while preserving candidate meanings. Models receive the complete policy, state and candidates, without the label, root identifier or derived feasibility hints.

For example, an otherwise eligible switch has action target \emph{switch} and permission target \emph{allow}. Changing only backup readiness to false changes these targets to \emph{wait} and \emph{block}. Changing only the answer-slot order must preserve the correct meaning. Together, the two changes test sensitivity to operational facts and resistance to superficial answer cues. The authored energy and time ranges place examples on both sides of the operational decision boundaries.

\subsection{Boundary cases and a worked example}
The nominal generator samples a down primary with valid evidence and a ready backup. Maximum evidence age is selected from $\{12,16,20,24,30\}$~s, switch duration from 4--15~s, completion margin from 2--6~s, and required recovery evidence from 2--4 distinct timestamps. The authored policy energy threshold spans 18--42 points and residual reserve 6--12 points. Nominal contact and energy values satisfy the corresponding constraints before a controlled transformation is applied. These ranges test whether a model follows the thresholds supplied in each policy.

The transformations make individual requirements consequential. Backup changes set readiness to false. Residual-energy cases set $E=E_R+c_E-1$, while policy-energy cases set $E=E_{\min}-1$. Closing-contact cases set $t_e=t+\ell+D+\delta-0.25$~s. Expiry cases set $t_m=t-\tau+0.25$~s, so evidence expires during the declared 0.5~s decision interval. Recovery changes invalidate or duplicate observations, remove an observation, mark the intervention unexecuted or change its completion time. Targets follow the complete priority rules after each change; an added fault does not necessarily change the answer when collecting fresh evidence already takes priority.

The released construction example illustrates how labels and answer slots are obtained. The example is a v0.3.0 test root selected to illustrate the backup transformation. At $t=30{,}371$~s, valid evidence from $30{,}361$~s has maximum age 12~s; contact is $[30{,}351,30{,}390)$~s. A switch takes 7~s with a 6~s margin. Energy is 44 points, switch cost 5, reserve 10 and policy threshold 38. With $\ell=0.5$~s, evidence age is 10.5~s, residual energy is 39 points and switching completes with its margin at $30{,}384.5$~s. Thus a ready backup makes \emph{switch} correct; changing only readiness to false makes \emph{wait} correct.

For A=collect, B=switch, C=monitor and D=wait, those targets are B/D. Rotating candidate meanings to A=switch, B=monitor, C=wait and D=collect changes the target letters to A/C. This separates the required change in operational meaning from a change in its answer slot. The model must use the supplied descriptions to identify the current answer slot for each operational meaning.

\subsection{Splits and suitability for the interface}
\begin{table}[t]
\caption{\normalfont Benchmark partitions.}
\label{tab:focus-protocol}
\centering\footnotesize
\setlength{\tabcolsep}{4pt}
\sbox{\satopstablebox}{%
\begin{tabular*}{.8\columnwidth}{@{\extracolsep{\fill}}lrr@{}}
\toprule
Partition & Questions & Roots \\
\midrule
Training & 1,536 & 1,536 \\
Development & 256 & 64 \\
Standard test & 1,024 & 256 \\
\midrule
Single failure & 384 & 96 \\
Compound failure & 384 & 96 \\
Recovery timing & 256 & 64 \\
Operator notes & 1,024 & 256 \\
\bottomrule
\end{tabular*}%
}%
\typeout{SATOPS-TABLE focus-protocol: width=\the\wd\satopstablebox; available=\the\linewidth}
\begin{minipage}{\wd\satopstablebox}
\usebox{\satopstablebox}
\par\smallskip{\raggedright\footnotesize Rows below the rule are diagnostic panels. Single/compound panels share 96 roots; operator notes re-render the 256 standard-test roots.\par}
\end{minipage}
\end{table}
The fixed training and development sets contain 1,536 and 256 questions for supervised updates and checkpoint selection, respectively. The main confirmation test contains 1,024 questions, 256 per task, from 256 independent roots. The 1,024 questions are grouped by 256 independent operational roots; each root contributes two related states and two renderings. Confirmation roots and normalized states are disjoint from training, development and earlier tests. They are newly sampled from the same authored grammar. Earlier pilot results motivated the adaptation study; its protocol was fixed before the reported domain updates and new confirmation inference.

Every main model is scored on the same standard test inputs and labels, in one fixed order. Accuracy is task-macro averaged; balanced task counts make it equal to overall question accuracy here. The proposed result is the mean and sample standard deviation across three fixed training seeds; each other model row reports one fixed configuration. A task-stratified paired root bootstrap keeps each root's correlated questions together. Auxiliary stress and recovery-timing panels probe the limits of the method. Single- and compound-failure panels share 96 roots and 192 nominal inputs; across the standard, two stress and recovery-timing panels, 2,048 questions come from 416 distinct roots and 1,856 distinct public inputs. Operator notes re-render the same 256 standard roots without adding roots or labels. These diagnostics are separate from the main standard accuracy comparison.

This construction suits Jev-style models because policy, state, question and a changing set of candidate meanings form a common decision interface across the four tasks. Qwen receives the same information and candidate meanings as the decision models. Independent contract implementations generate the gold labels, and exact canonical rules reach 100\%. A held-out operator-note rendering preserves the same facts in sentences, testing whether structured-input gains transfer beyond the training representation. The benchmark evaluates the common decision interface and the authored operational logic.
\section{Evaluation}
\subsection{Models and measurement}
\emph{SatOps-Decision (ours)} denotes our domain-adapted local Laya scorer. Raw accuracy scores its answers before the separate operational checker, and latency measures its complete model request. Local comparators are frozen Laya, generative Qwen3.5-0.8B and Qwen-derived OneJev1.6-0.8B. Original TypeSafe Jev 1.13/1.14 provide cloud accuracy references through the API. Qwen generates a short answer greedily with thinking disabled and a 16-token cap; malformed outputs remain incorrect. All local inputs are checked for truncation. Original Jev was added after the local results were fixed, receives the same 1,024 standard questions, and supplies only raw accuracy.

We measure complete-request latency for all locally deployed models on the same computer. The setup uses an Apple M6 with 24~GiB memory, macOS~27.0, PyTorch MPS, FP32 and four host threads. After accuracy evaluation, a preselected balanced subset of 128 standard-test questions is timed over three shuffled passes after eight warmup requests, with model order varied, giving 384 timed requests per model. Complete-request time includes formatting, tokenization, transfer, synchronized inference and answer extraction; model loading, file I/O and the separate execution checker are excluded. Timing responses must reproduce the accuracy pass. Seed~7 is predesignated for proposed-model timing.

\subsection{Primary benefit: short local request latency}
\begin{table}[t]
\caption{\normalfont Local decision latency.}
\label{tab:focus-runtime}
\centering\footnotesize
\setlength{\tabcolsep}{4pt}
\sbox{\satopstablebox}{%
\begin{tabular*}{.8\columnwidth}{@{\extracolsep{\fill}}lrr@{}}
\toprule
Model & \shortstack{Median\\(ms/decision)} & \shortstack{P95\\(ms/decision)} \\
\midrule
\textbf{SatOps-Decision (ours)} & \textbf{90.3} & 104.7 \\
Laya (frozen) & 91.2 & \textbf{104.5} \\
Qwen3.5-0.8B & 493.9 & 507.8 \\
OneJev1.6-0.8B & 388.4 & 403.0 \\
\bottomrule
\end{tabular*}%
}%
\typeout{SATOPS-TABLE focus-runtime: width=\the\wd\satopstablebox; available=\the\linewidth}
\begin{minipage}{\wd\satopstablebox}
\usebox{\satopstablebox}
\par\smallskip{\raggedright\footnotesize 384 requests per model; ours uses seed 7. Bold: lowest latency in each column.\par}
\end{minipage}
\end{table}
SatOps-Decision's local scorer has a median complete-request latency of 90.3~ms, compared with 493.9~ms for Qwen generation and 388.4~ms for OneJev. These are \textbf{5.47$\times$} and \textbf{4.30$\times$} control-to-proposed time ratios on the same inputs and local execution setup, calculated from unrounded measurements. Frozen Laya takes 91.2~ms. Domain adaptation raises mean raw standard-test accuracy from 42.5\% to 85.3\% while retaining approximately the same measured inference cost. Across all adapted seeds, medians range from 90.1 to 90.3~ms.

Direct candidate scoring removes the answer-decoding stage from each request. The timings characterize the evaluated checkpoints and implementations on the stated desktop FP32/MPS setup. The decoder controls use reference kernels with shared-prefix caching disabled. Original Jev API duration combines provider inference with network and service overhead, so its evaluation supplies an accuracy reference.

Shorter requests preserve more evidence freshness and contact time in~\eqref{eq:permit}, motivating direct decisions in the operations path. Table~\ref{tab:focus-runtime} measures one request; sequential action and permission queries accumulate their durations before dispatch. With independent-request batching, answers become available only when the batch finishes, while amortized time per question describes throughput. Our separate batch diagnostic showed no benefit for this implementation. Median and P95 summarize typical and upper-tail measured request times. Dispatch also depends on readiness and energy, while recovery is verified from subsequent observations. Relative to the declared 0.5~s interval, one median request consumes 18.1\% (ours) or 98.8\% (Qwen) of the budget; these fractions describe computation headroom.

\subsection{Decision accuracy after domain adaptation}
\begin{table}[t]
\caption{\normalfont Accuracy on the same 1,024 test questions.}
\label{tab:focus-accuracy}
\centering\footnotesize
\setlength{\tabcolsep}{4pt}
\sbox{\satopstablebox}{%
\begin{tabular*}{.8\columnwidth}{@{\extracolsep{\fill}}llr@{}}
\toprule
Model & Execution & Accuracy (\%) \\
\midrule
\textbf{SatOps-Decision (ours)} & Local & 85.3 $\pm$ 0.8 \\
Laya (frozen) & Local & 42.5 \\
Qwen3.5-0.8B & Local & 40.7 \\
OneJev1.6-0.8B & Local & 58.6 \\
Jev 1.14 & Cloud API & \textbf{87.4} \\
Jev 1.13 & Cloud API & 81.9 \\
\bottomrule
\end{tabular*}%
}%
\typeout{SATOPS-TABLE focus-accuracy: width=\the\wd\satopstablebox; available=\the\linewidth}
\begin{minipage}{\wd\satopstablebox}
\usebox{\satopstablebox}
\par\smallskip{\raggedright\footnotesize Ours: mean $\pm$ SD. Bold value: highest accuracy.\par}
\end{minipage}
\end{table}
Table~\ref{tab:focus-accuracy} evaluates five model categories on identical held-out questions. SatOps-Decision reaches 85.3$\pm$0.8\% raw accuracy, compared with 42.5\% frozen Laya. The gain is 42.77 percentage points (paired root-bootstrap 95\% interval [40.14, 45.15]). The three seeds answer 865, 882 and 872 of 1,024 questions correctly. Qwen generation and OneJev reach 40.7\% and 58.6\%. Original Jev 1.14 answers 895 questions correctly (87.4\%); Jev 1.13 answers 839 (81.9\%). The paired interval quantifies the adaptation gain over frozen Laya.
The comparison with frozen Laya isolates the benefit of supervised domain adaptation within the same base model and interface. The other evaluated models differ in pretraining, size and adaptation. Local Laya provides an adaptable, low-latency component for the proposed operations method.

On 256 questions per task, adapted accuracy is 74.5\% for assessment, 89.1\% for action, 78.9\% for permission and 98.6\% for recovery; frozen Laya obtains 37.1\%, 25.8\%, 53.9\% and 53.1\%, respectively. Adaptation improves every evaluated stage, with its largest gain in action selection and its lowest resulting accuracy in assessment. Assessment and action have four candidate meanings; permission and recovery are binary. All stages use the same selected checkpoint and native Choice interface.

\subsection{What changes through adaptation}
\begin{table}[t]
\caption{\normalfont Adaptation controls on the standard test (seed 7).}
\label{tab:conference-ablation}
\centering\footnotesize
\setlength{\tabcolsep}{4pt}
\begin{tabular*}{.8\columnwidth}{@{\extracolsep{\fill}}lrr@{}}
\toprule
Training configuration & \shortstack{Updated\\parameters (M)} & \shortstack{Raw accuracy\\(\%)} \\
\midrule
\textbf{Full adaptation (ours)} & 421.03 & 84.47 \\
Head only & 26.25 & 45.70 \\
No order augmentation & 421.03 & \textbf{84.67} \\
\bottomrule
\end{tabular*}
\end{table}

Table~\ref{tab:conference-ablation} reports the prescribed single-seed controls, each selected on development data and evaluated on the same 1,024 confirmation questions. The head-only run freezes and detaches the encoder; the no-augmentation run retains full adaptation but keeps native candidate appearance fixed during training. Full adaptation answers 865 questions correctly, versus 468 for head-only adaptation. This supports joint adaptation of the encoder and decision layers for the authored domain. The no-augmentation run answers 867 correctly (84.67\%), compared with 865 (84.47\%) for full adaptation under the same seed.

The main comparison measures the learned scorer before operational checking. Each adapted seed makes 54 false authorizations among 128 gold-ineligible permission questions (42.2\%), with no eligible switches blocked. All 28 action errors are unwarranted switches among 192 questions whose correct action is not switch. Adding deterministic corrections raises the checked system's mean standard accuracy to 93.3\%; the main accuracy table reports the learned scorer before these corrections. These directional errors motivate checking each proposed switch against the operational contract before dispatch.

\subsection{Input transfer and recovery diagnostics}
Canonical rules solve the fully specified contract exactly. On matched operator-note inputs, adapted Laya reaches 47.4\%, compared with OneJev's 56.9\%. In the 24-case clocked event diagnostic, checked adapted Laya restores 4--5 cases per seed, compared with 8 for checked frozen Laya. Because the checker preserves negative proposals, successful recovery depends on the scorer proposing a useful action. These diagnostics identify input-representation transfer and recovery decisions as priorities for further development. The full report provides the stress, ablation and directional-error results.
\section{Conclusion}
We presented SatOps-Decision for low-latency decisions in onboard satellite communication operations. To the best of our knowledge, its two contributions are the first application of Jev-style general-purpose discriminative decision models to satellite operations and the first dedicated benchmark for this approach. A domain-adapted local Laya scorer implements the method; independently checked labels support the benchmark. Evaluation prioritizes request latency and raw held-out accuracy. On the measured FP32/MPS implementation, SatOps-Decision's scorer has a median complete-request latency of 90.3~ms, a 5.47$\times$ speedup over Qwen generation, while adaptation raises raw standard-test accuracy from 42.5\% to 85.3$\pm$0.8\%. Original Jev 1.14 reaches 87.4\% on the same questions as an accuracy-only cloud reference. Next steps include improving input-representation transfer and recovery decisions, followed by validation with real telemetry, mission-specific constraints and onboard hardware.

\balance
\bibliographystyle{IEEEtran}
\bibliography{satops-references-34eb0b31b2b9}
}
\begin{document}
\SatOpsManuscript
\end{document}